\documentclass{article}
\usepackage{spconf,amsmath,graphicx,hyperref}
\hypersetup{hidelinks}
\usepackage{xcolor}

\definecolor{myblue}{RGB}{0,70,140}

\hypersetup{
    colorlinks=true,
    linkcolor=myblue,
    citecolor=myblue,
    urlcolor=myblue
}
\usepackage{booktabs}
\usepackage{subcaption}
\usepackage{graphicx}
\usepackage{multirow}
\usepackage{xcolor}
\usepackage{color, colortbl}
\usepackage{xspace}
\makeatletter
\DeclareRobustCommand\onedot{\futurelet\@let@token\@onedot}
\def\@onedot{\ifx\@let@token.\else.\null\fi\xspace}

\makeatother

\newcommand{\parhead}[1]{\noindent\textbf{#1}\xspace}
\usepackage{multirow}%
\usepackage{algorithmic}
\usepackage{graphicx}
\usepackage{booktabs}
\usepackage{textcomp}
\usepackage{tikz}
\usepackage{xcolor}
\newcommand{\ourmethod}{\textit{ViCo-SAM3}}
\definecolor{myPink}{RGB}{255,226,237}
\definecolor{myTeal}{RGB}{0,161,106}
\definecolor{myOrange}{RGB}{249,70,0}
\definecolor{myOrange1}{RGB}{242,170,132}
\definecolor{nature}{RGB}{101, 164, 135}
\definecolor{medical}{RGB}{104, 52, 154}

\definecolor{myBule}{RGB}{0,104,211}
\definecolor{reda}{RGB}{202,0,0}
\definecolor{mygreen}{RGB}{0,136,51}
\definecolor{myOrange2}{RGB}{197, 90, 17}
\definecolor{myzishe}{HTML}{E6E6FA}
\definecolor{Light}{rgb}{0.92, 0.99, 0.95}
\definecolor{color1}{HTML}{ECF4F9}
\definecolor{color2}{HTML}{FFF1E0}
\definecolor{color3}{HTML}{ECF4E9}
\usepackage{cleveref}
\crefname{section}{\S}{\S\S}
\Crefname{section}{\S}{\S\S}
\crefname{figure}{fig.}{figs.}
\Crefname{figure}{Fig.}{Figs.}
\crefname{table}{tab.}{tabs.}
\Crefname{table}{Tab.}{Tabs.}
\crefname{equation}{eq.}{eqs.}
\Crefname{equation}{Eq.}{Eqs.}
\crefname{algorithm}{alg.}{algs.}
\Crefname{algorithm}{Alg.}{Algs.}
\usepackage{enumitem}
\begin{document}
%\title{\ourmethod: Adapting SAM 3 for Open-Vocabulary Camouflaged Object Segmentation via Vision-Conditioned Context\\}
\title{\ourmethod: Vision-Conditioned Alignment for Open-Vocabulary Camouflaged Object Segmentation\\}
\name{Qiangqiang Zhou$^{1}$, Wenjun Tang*$^{1}$\thanks{* Corresponding author: \{twj1010, jiawei\_xu\}@jxnu.edu.cn}, Yong Chen$^{1}$, Dandan Zhu$^{2}$, Jiawei Xu$^{1}$*}
\address{$^{1}$School of Artificial Intelligence, Jiangxi Normal University\\
$^{2}$Institute of AI Education, East China Normal University}
\maketitle

\begin{abstract}
Open‑vocabulary camouflaged object segmentation (OVCOS) aims to segment unseen camouflaged objects under text guidance. 
%Existing VLM‑based methods suffer from a semantic gap between global text descriptions and pixel‑level cues, leading to confusion and prompt failure. 
We observe that SAM3 still suffers from a pronounced semantic gap between global textual semantics and fine-grained pixel-level visual cues in OVCOS. Meanwhile, fully fine-tuning the text encoder introduces heavy parameter overhead and risks overfitting to training categories, which compromises open-vocabulary representation flexibility. 
To address these issues, we propose \ourmethod, a \underline{\textbf{Vi}}sion-\underline{\textbf{Co}}nditioned alignment framework designed for OVCOS. Specifically, we introduce vision-conditioned (ViCo) module, which dynamically modulates text embeddings with global visual context, enabling textual representations to adapt to the current image content and thereby effectively bridging the semantic gap between vision and text.
Building on this, we further design a vision-conditioned cross-modal binding (ViCoBind) module to enhance cross-modal interaction and semantic alignment between visual and textual representations.
%Full fine‑tuning of encoders exacerbates this by causing catastrophic forgetting of rare categories. To address these issues, we introduce a Vision‑Conditioned Dynamic Text Adapter on SAM3 that extracts global visual context to dynamically reshape text embeddings. 
%
%By injecting background priors into textual anchors, our approach creates discriminative cross‑modal anomalies for precise object binding without altering pre‑trained features. Extensive experiments on OVCamo demonstrate new state‑of‑the‑art performance (cIoU 0.773) and strong zero‑shot generalization, effectively overcoming semantic drift.
Without bells and whistles, \ourmethod ~achieves state-of-the-art performance on the OVCamo benchmark and demonstrates strong generalization.
\end{abstract}


\section{Introduction}
%Camouflaged Object Segmentation (COS) aims to identify objects that seamlessly blend into their surroundings. While deep learning has advanced this field, traditional models operate under a closed set assumption, failing to generalize to unseen categories \cite{b2}\cite{b3}. 
%
%To overcome this, Open‑Vocabulary Camouflaged Object Segmentation (OVCOS) has recently emerged, leveraging natural language prompts to segment novel camouflaged objects \cite{b1}. 
%Open‑Vocabulary Camouflaged Object Segmentation (OVCOS) has recently emerged, leveraging natural language prompts to segment novel camouflaged objects. 
%Current OVCOS methods often build upon Vision‑Language Models (VLMs) like and foundation segmentation models. However, most adopt a unidirectional text to vision paradigm, where static textual embeddings guide visual feature matching. In highly ambiguous camouflage scenes, this one way interaction causes severe semantic confusion the model struggles to distinguish the target from visually similar distractors. 
%
Open-vocabulary camouflaged object segmentation (OVCOS)~\cite{b1,b13,zhao2026open} has recently emerged as a challenging task that aims to segment unseen camouflaged objects under natural language guidance. Recent foundation models~\cite{b4,siglip,grounding} such as SAM3~\cite{SAM3} provide strong open-vocabulary segmentation capabilities~\cite{b10,li2022language}, yet directly applying them to OVCOS remains non-trivial. We observe a pronounced semantic gap between global textual semantics and the fine-grained visual evidence required to distinguish camouflaged objects~\cite{b1,b13}. In highly ambiguous camouflage scenes, static text representations cannot adapt to the surrounding visual context, making it difficult to establish precise semantic correspondence between the prompted concept and visually similar foreground-background regions.
%
%A straightforward solution is to fully fine-tune the encoder on camouflage data. However, such domain-specific adaptation may distort the semantic space inherited from large-scale pre-training, compromising open-vocabulary generalization and causing catastrophic forgetting, particularly for rare or unseen categories.

\begin{figure}[t]
    \centering
    \includegraphics[width=0.9\linewidth]{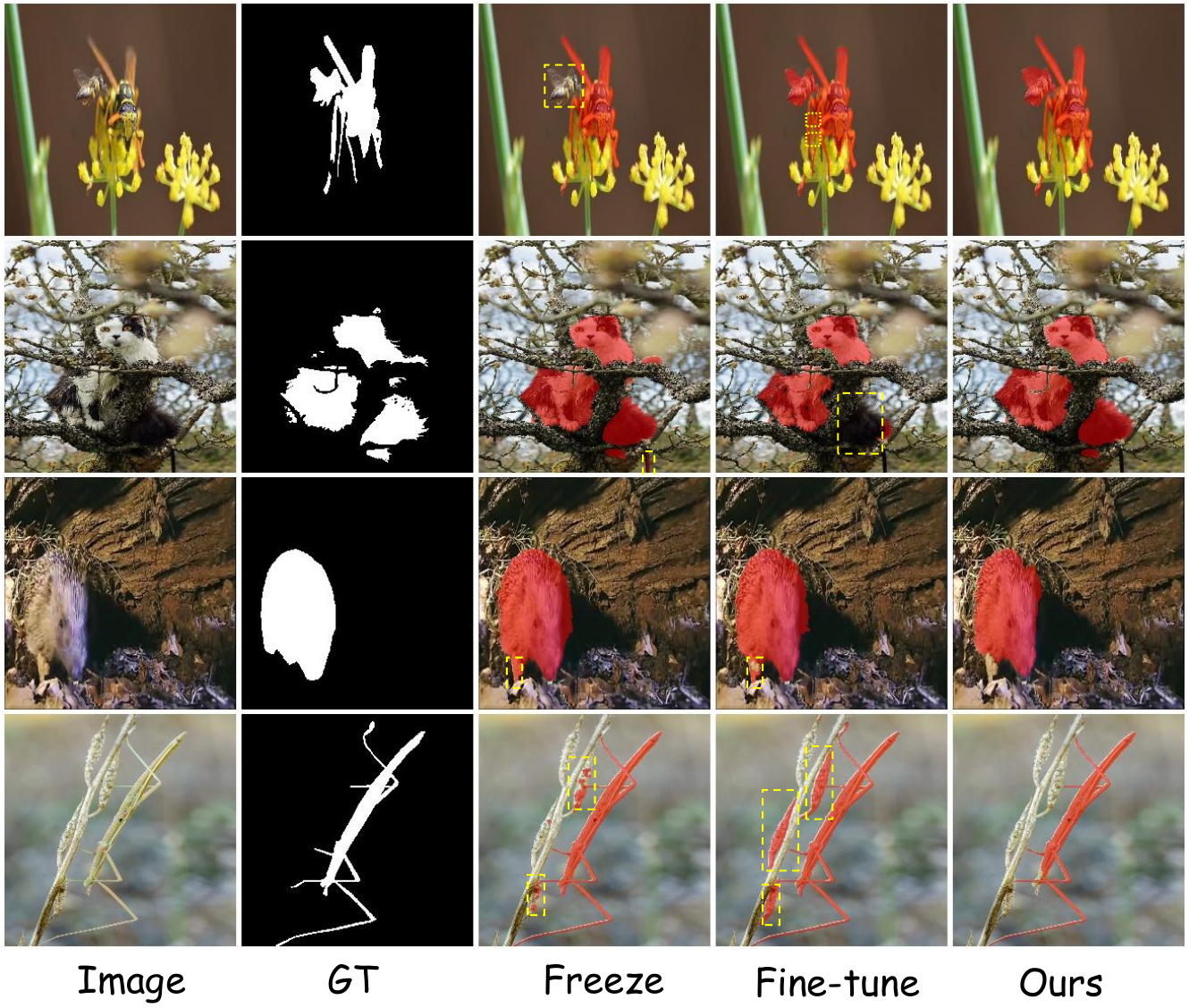}
   \caption{Qualitative comparison of different text adaptation strategies. From left to right: input image, GT, frozen text encoder, fully fine-tuned text encoder, and our method.}
    \label{fig:image1}
\end{figure}
A straightforward solution is to fully fine-tune the encoder on camouflage data. However,as shown in \Cref{fig:image1}, such domain-specific adaptation may alter the general semantic representations acquired during large-scale pre-training, leading to catastrophic forgetting of pre-trained knowledge and further compromising open-vocabulary generalization, particularly for rare and unseen categories.
%
%This presents a fundamental dilemma for OVCOS: the model needs to adapt its semantic representation to highly ambiguous camouflage scenes while preserving the rich open-vocabulary knowledge inherited from large-scale pre-training.
This creates a fundamental trade-off in OVCOS between adapting semantic representations to ambiguous camouflage scenes and preserving the open-vocabulary knowledge inherited from large-scale pre-training.

\begin{figure*}[t]
    \centering
    \includegraphics[width=0.9\linewidth]{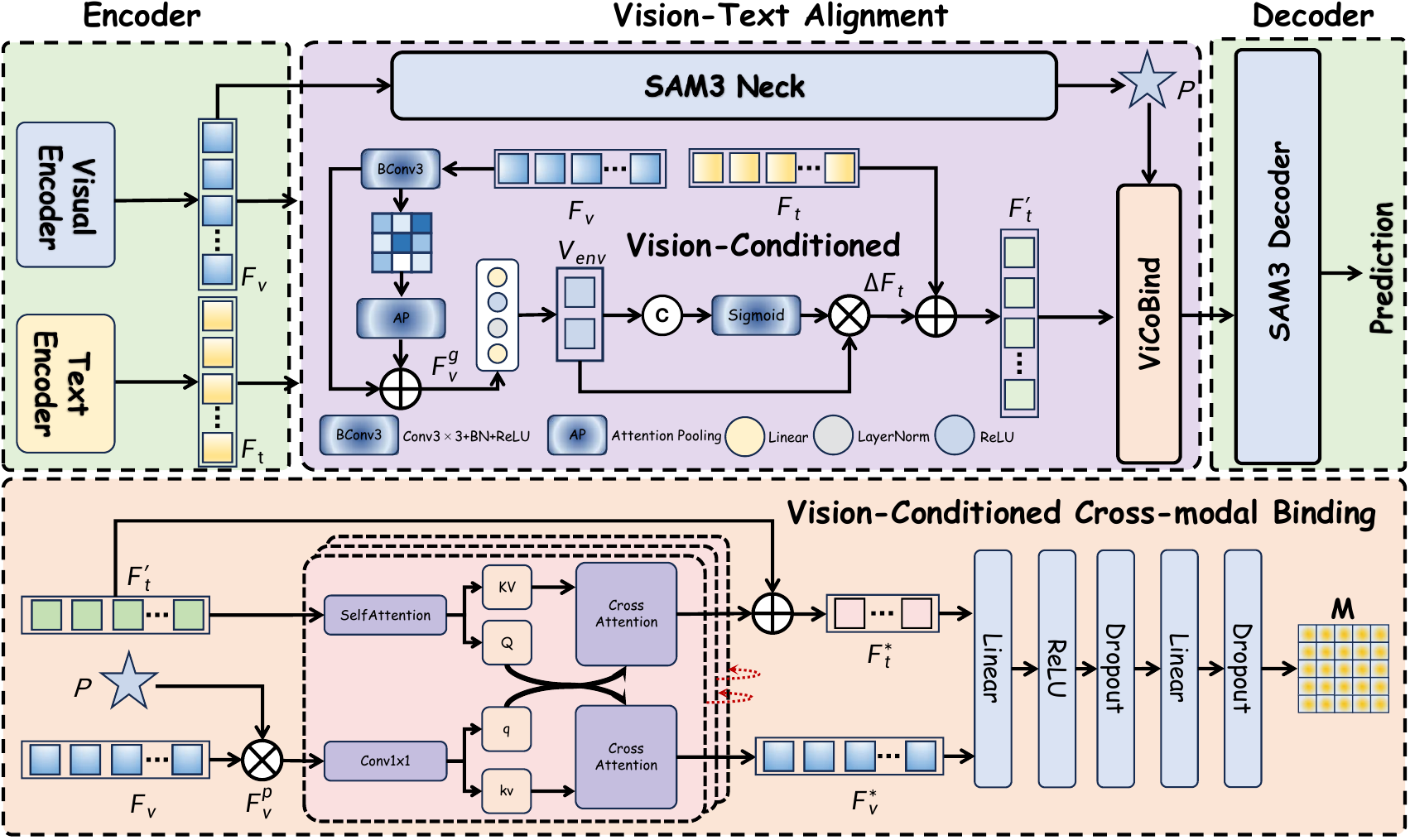}
    \caption{Overall pipeline of \ourmethod. To enhance vision-text alignment, we introduce vision-conditioned (ViCo) module and Vision-Conditioned Cross-modal Binding (ViCoBind) module between the SAM3 encoders and decoder, enabling effective adaptation and interaction between visual and textual representations.}
    \label{fig:image2}
\end{figure*}
%A naive remedy is to fully finetune the VLM encoders on camouflage data, but this induces a semantic drift bottleneck: domain‑specific adaptation improves in domain accuracy at the cost of corrupting the pretrained semantic space, leading to catastrophic forgetting on rare open‑vocabulary categories.

%To address this dilemma, we propose \ourmethod (Vision‑Conditioned Dynamic Text Adaptation SAM3), a novel framework built upon SAM3 for OVCOS. Unlike static text embeddings, our approach dynamically reshapes textual anchors conditioned on global visual context. 
To address these issues, we propose \textit{\ourmethod}, a vision-conditioned alignment framework built upon SAM3 for OVCOS. Rather than modifying the pre-trained text encoder, \ourmethod ~adapts textual representations to the current visual context while preserving the original open-vocabulary representation space. Specifically, we introduce a vision-conditioned (ViCo) module that leverages global visual context to dynamically modulate text embeddings, thereby bridging the semantic gap between global textual semantics and fine-grained visual cues.
Building on this, we further introduce a vision-conditioned cross-modal binding (ViCoBind) module that facilitates bidirectional information exchange between visual and textual features, further strengthening cross-modal interaction and semantic alignment. Together, these modules enable more precise semantic binding of camouflaged objects while preserving the model's open-vocabulary generalization capability.
%
%By injecting environmental visual priors into the text branch, we transform the camouflaged object into a distinct cross‑modal anomaly, enabling precise object binding while preserving the pre‑trained open‑vocabulary knowledge.

%Extensive experiments on the OVCamo benchmark demonstrate that \ourmethod ~escapes the semantic drift trap, achieving new state‑of‑the‑art performance across all metrics with robust zero‑shot generalization on challenging unseen categories.
Extensive experiments on the OVCamo benchmark demonstrate that \ourmethod ~effectively mitigates semantic drift, achieves
state-of-the-art performance on the OVCamo benchmark and
demonstrates strong generalization to unseen categories.
\textit{Our contributions are summarized as follows:}
\begin{itemize}
[leftmargin=*,itemsep=0em,topsep=0em,parsep=0em]
%\item
%We identify a semantic gap between textual semantics and fine-grained visual cues in OVCOS, as well as the risk of catastrophic forgetting from text-encoder fine-tuning.
%\item We introduce ViCo, which dynamically modulates text embeddings using global visual context, and further design a bidirectional cross-modal interaction module to enhance semantic alignment between vision and language.
%\item Experiments on the OVCamo benchmark demonstrate that \textit{\ourmethod} achieves new state-of-the-art performance and exhibits strong zero-shot generalization.
\item
We identify a semantic gap between textual semantics and fine-grained visual cues in OVCOS, as well as the risk of catastrophic forgetting from text-encoder fine-tuning, and propose \ourmethod ~to address these challenges.

\item
We introduce ViCo, which dynamically adapts text embeddings using global visual context while preserving pre-trained open-vocabulary knowledge.

\item
We design ViCoBind, a bidirectional cross-modal binding module that strengthens interaction and semantic alignment between visual and textual features.

\item
%\ourmethod ~achieves state-of-the-art performance on the OVCamo benchmark and demonstrates strong zero-shot generalization.
\ourmethod ~achieves state-of-the-art performance on OVCamo with strong generalization.
\end{itemize}

\section{Methodology}
\parhead{Overall Architecture.}
%\subsection{Overall Architecture}
%\ourmethod ~consists of two components: \textit{the visual stream} and \textit{the text stream}.
As shown in \Cref{fig:image2}, we first employ the visual and frozen text encoders of SAM3 to extract the visual feature $F_v$ and text feature $F_t$ from the input image and text prompt, respectively. The resulting $F_v$ and $F_t$ are then fed into the vision-conditioned (ViCo) module, where the frozen textual representation is dynamically aligned with the visual context to obtain the vision-adaptive text feature $F_t'$. Subsequently, $F_t'$, the $F_v$, and the positional feature $P$ derived from $F_v$ through the SAM3 Neck are further aligned and fused in the ViCoBind module, producing the fused representation $M$. Finally, $M$ is fed into the SAM3 decoder to generate the final segmentation prediction.
\parhead{Vision-Conditioned (ViCo) Module.}
%\subsection{Vision-Conditioned (ViCo) Module}
To bridge the pronounced semantic gap between global textual semantics and fine-grained pixel-level visual cues, we introduce a ViCo to explicitly model the interaction between visual context and textual semantics. 
%
%Given the visual feature $F_v$, we first enhance its local representation using a BConv $3\times3$ block, composed of a $3\times3$ convolution, BN, and ReLU activation. The enhanced feature is then processed by Attention Pooling to aggregate global contextual information:
%\begin{equation}
%    \hat{F}_v = \mathrm{BConv}~{3\times3}(F_v), \qquad
%    z_v = \mathrm{AP}(\hat{F}_v).
%\end{equation}
%where $\mathrm{BConv}~{3\times3}$ consists of a $3\times3$ convolution, BN, and ReLU.
Given the visual feature $F_v$, we first enhance its local representation using a BConv$3\times3$ block, which consists of a $3\times3$ convolution, Batch Normalization (BN), and ReLU activation. Attention Pooling (AP) is then applied to aggregate global visual context:
\begin{equation}
    F_v^{g}
    =
    \mathrm{AP}\left(
    \mathrm{BConv 3\times3}(F_v)
    \right),
\end{equation}
where $F_v^{g}$ denotes the aggregated global visual representation.
The pooled representation is then combined with a residual visual branch and projected into a compact environmental representation:
\begin{equation}
    V_{\mathrm{env}}
    = \phi_e\big(F_v^{g} + \mathcal{R}(F_v)\big),
\end{equation}
where $\mathcal{R}(\cdot)$ denotes the residual visual projection and $\phi_e(\cdot)$ is a lightweight linear mapping. $V_{\mathrm{env}}$ therefore summarizes the global visual context of the current image.
We further use $V_{\mathrm{env}}$ to dynamically adapt the original text feature $F_t$. Specifically, $F_t$ and $V_{\mathrm{env}}$ are concatenated to estimate a vision-conditioned gating weight ($G$):
\begin{equation}
    G = \sigma\left(
    \phi_g\big([F_t;V_{\mathrm{env}}]\big)
    \right),
\end{equation}
where $[\cdot;\cdot]$ denotes feature concatenation, $\phi_g(\cdot)$ is a learnable projection, and $\sigma(\cdot)$ denotes the Sigmoid function. The resulting gate controls the contribution of the visual context to the textual representation, yielding the semantic offset :
\begin{equation}
    \Delta F_t = G \odot V_{\mathrm{env}},
\end{equation}
where $\odot$ denotes element-wise multiplication. Finally, the offset is injected into the original text feature through a residual connection:
\begin{equation}
    F_t' = F_t + \Delta F_t.
\end{equation}
%In this way, ViCo preserves the original semantic knowledge encoded in $F_t$ while dynamically adapting it to the visual context of each input image, thereby improving the alignment between textual semantics and fine-grained visual cues.

%\subsection{Vision-Conditioned Cross-modal Binding Module}
\parhead{Vision-Conditioned Cross-modal Binding Module.}
To further enhance cross-modal interaction and semantic alignment, we introduce the ViCoBind, which establishes fine-grained bidirectional interactions between the vision-adaptive text representation $F_t'$ and visual features $F_v$.
Specifically, we first inject the positional prior $P$ into the visual feature $F_v$ to obtain a position-aware visual representation:
\begin{equation}
    F_v^{p} = F_v \odot P,
\end{equation}
where $\odot$ denotes element-wise multiplication.
Meanwhile, the adapted text feature $F_t'$ is refined through self-attention, while $F_v^{p}$ is projected by a $1\times1$ convolution:
\begin{equation}
    Q_t, K_t, V_t = \mathrm{SelfAttn}(F_t'),
\end{equation}
\begin{equation}
    q_v, k_v, v_v = \mathrm{Conv}_{1\times1}(F_v^{p}).
\end{equation}
We then perform bidirectional cross-modal attention. On the one hand, visual queries attend to textual keys and values to inject semantic guidance into each spatial location. On the other hand, textual queries attend to visual keys and values, allowing the textual representation to capture image-specific visual evidence:
\begin{equation}
\left\{
\begin{aligned}
\hat{F}_v =
    \mathrm{Softmax}
    \left(
    \frac{q_v K_t^{\top}}{\sqrt{d}}
    \right)V_t,\\
    \hat{F}_t =
    \mathrm{Softmax}
    \left(
    \frac{Q_t k_v^{\top}}{\sqrt{d}}
    \right)v_v.
\end{aligned}
\right.
\end{equation}

Residual connections are further employed to preserve the original unimodal information:
\begin{equation}
    F_v^{*} = F_v^{p} + \hat{F}_v,
    \qquad
    F_t^{*} = F_t' + \hat{F}_t.
\end{equation}
Such bidirectional interaction is performed iteratively for $L$ layers, enabling progressive information exchange between textual semantics and spatial visual cues.

Finally, the aligned representations $F_t^{*}$ and $F_v^{*}$ are jointly fused and passed through a lightweight FFN:
\begin{equation}
    M =
    \mathrm{FFN}\left(
    \mathrm{Fuse}(F_t^{*},F_v^{*})
    \right),
\end{equation}
The resulting representation $M$ is subsequently fed into the SAM3 decoder for mask prediction ($\hat{P}$).
Following previous methods~\cite{xu2025semantic,xu2026hvpnet,tpseg,differseg}, \ourmethod ~is trained with a combination of binary cross-entropy and weighted IoU losses.

\begin{table*}[t]
\centering
%\caption{Comparison with state-of-the-art open-vocabulary semantic image segmentation methods with different training settings on the OVCamo. Black Bold denote the best results. Notably, ‘None’ indicates that the model does not leverage an additional backbone (e.g., ResNet, Swin). }
\caption{Comparisons with state-of-the-art models on OVCamo dataset. We report the \colorbox{color1}{visual-text encoder}, \colorbox{color2}{auxiliary encoder}, and \colorbox{color3}{text prompt} used by each method. \textbf{Bold} indicate the best results.$^{\dagger}$ denotes the model evaluated in a zero-shot manner without any training on the OVCamo training set.}
\label{tab:tab1}
\resizebox{0.95\linewidth}{!}{
\begin{tabular}{r|c|c|c|cccccc}
\toprule
\multicolumn{1}{c|}{Model} & \cellcolor{color1}Visual-Text Encoder & \cellcolor{color2}Auxiliary Encoder & \cellcolor{color3}Text Prompt & $cS_m \uparrow$ & $cF_\omega \uparrow$ & $cMAE \downarrow$ & $cF_m \uparrow$ & $cE_m \uparrow$ & $cIoU \uparrow$ \\ 
\bottomrule
SimSeg \cite{b6} & CLIP-ViT-B/16~\cite{b4} & ResNet-101\cite{he2016deep} & Learnable~\cite{b14} & 0.053 & 0.049 & 0.921 & 0.056 & 0.098 & 0.047 \\
OVSeg \cite{b7}  & CLIP-ViT-L/14~\cite{b4} & Swin-B\cite{liu2021swin} & DefaultPrompts~\cite{b15} & 0.024 & 0.046 & 0.954 & 0.056 & 0.130 & 0.046 \\
SAN \cite{b9}    & CLIP-ViT-L/14~\cite{b4} & ViT Adapter~\cite{chen2022vision} & DefaultPrompts~\cite{b15} & 0.275 & 0.202 & 0.612 & 0.220 & 0.318 & 0.189 \\
CAT-Seg~\cite{b10} & CLIP-ViT-L/14~\cite{b4} & Swin-B~\cite{liu2021swin} & DefaultPrompts~\cite{b15} & 0.181 & 0.106 & 0.719 & 0.123 & 0.196 & 0.094 \\
SuCLIP~\cite{SuClip} & CLIP-ConvNeXt-L~\cite{liu2022convnet} & None & CAP~\cite{SuClip} & 0.533 & 0.449 & 0.368 & 0.482 & 0.570 & 0.395 \\
OVCoser~\cite{b1}& CLIP-ConvNeXt-L~\cite{liu2022convnet} & None & CamoPrompts~\cite{b1} & 0.579 & 0.490 & 0.336 & 0.520 & 0.616 & 0.443 \\
BaClip \cite{b13}& CLIP-ConvNeXt-L~\cite{liu2022convnet} & None & CamoPrompts~\cite{b1} & 0.589 & 0.540 & 0.327 & 0.559 & 0.640 & 0.488 \\
SAM3$^{\dagger}$~\cite{SAM3}& SAM3~\cite{SAM3} & None & CamoPrompts~\cite{b1} & 0.735 & 0.592 & 0.086 & 0.605 & 0.786 & 0.552\\
\rowcolor{myzishe}
\textbf{Ours} & SAM3~\cite{SAM3} & None & CamoPrompts~\cite{b1} & \textbf{0.889} & \textbf{0.838} & \textbf{0.015} & \textbf{0.859} & \textbf{0.947} & \textbf{0.773} \\
\bottomrule
\hline
\end{tabular}%
}
\vspace{-3mm}
\end{table*}

\section{Experimental}
\vspace{-1mm}
\subsection{Evaluation}
\vspace{-1mm}
\parhead{Implementation Details.}
%The \ourmethod ~is implemented using the PyTorch framework and is trained on a single NVIDIA GeForce RTX 4090 GPU. During training, each input image is resized to 384 × 384.%, with the SAM3 parameters remaining frozen. 
%We employ the AdamW optimizer with hyperparameters: batch size = 4, learning rate = $3 \times 10^{-5}$, epochs = 30. A cosine annealing scheduler adjusts the learning rate. Basic data augmentations, including random flipping, rotating, and color jittering, are introduced to preprocess training data.
\ourmethod ~is implemented in PyTorch and trained on a single NVIDIA GeForce RTX 4090 GPU. All input images are resized to $384 \times 384$. We use AdamW with a batch size of 4 and an initial learning rate of $3 \times 10^{-5}$ for 30 epochs.%, together with a cosine annealing learning rate schedule. Standard data augmentations, including random flipping, rotation, and color jittering, are applied during training.

\parhead{Datasets and Metrics.}
 %We evaluate \ourmethod ~on the standard OVCOS benchmark, OVCamo. To rigorously test the zero-shot generalization capabilities on unseen and rare species, our evaluation is conducted across the complete set of 3,770 highly ambiguous test samples. Top-1 classification accuracy is reported on correctly identified camouflaged regions. For segmentation evaluation, standard metrics including $cIOU$ (classaware Intersection over Union), $cS_{m}$ (class-aware S-measure), $cMAE$ (class-aware Mean Absolute Error), $cF_{\beta}$ (class-aware F-measure), $cF_{\beta}^{\omega}$ (class-aware weighted F-measure), and $cE_{m}$ (class-aware E-measure) are employed to comprehensively assess segmentation performance.
 %We evaluate \ourmethod ~on the OVCamo benchmark, using all 3,770 test samples to assess zero-shot generalization to unseen and rare categories. We report six class-aware segmentation metrics \cite{b1}: $cIoU$, $cS_m$, $cMAE$, $cF_{m}$, $cF_{\omega}$, and $cE_m$.
 We evaluate \ourmethod ~on the OVCamo~\cite{b1} benchmark using all 3,770 test samples, which span 61 categories that are completely unseen during training, thereby enabling a strict evaluation of zero-shot generalization. We report six class-aware segmentation metrics~\cite{b1}: $cIoU$, $cS_m$, $cMAE$, $cF_{m}$, $cF_{\omega}$, and $cE_m$.
%\subsection{Comparison with SOTA}

\parhead{Quantitative Comparison.} 
%We compare our \ourmethod ~with recent state-of-the-art methods~\cite{OVCoser,SuClip}. The quantitative results on the OVCamo benchmark are summarized in \Cref{tab:tab1}. \ourmethod ~consistently outperforms all competing methods across all metrics. Notably, our method achieves a significant performance margin in terms of $cF_{\beta}$ and $cE_{m}$, indicating that the vision-conditioned dynamic text adaptation effectively mitigates semantic confusion and provides vastly superior pixel-level alignment.
As shown in \Cref{tab:tab1}, \ourmethod ~achieves the best results across all metrics on OVCamo. The notable improvements in $cF_{m}$ and $cE_m$ indicate stronger region discrimination and structural alignment, validating the effectiveness of vision-conditioned dynamic text adaptation in reducing semantic ambiguity.

\begin{figure}[t] 
    \centering
    \includegraphics[width=1\linewidth]{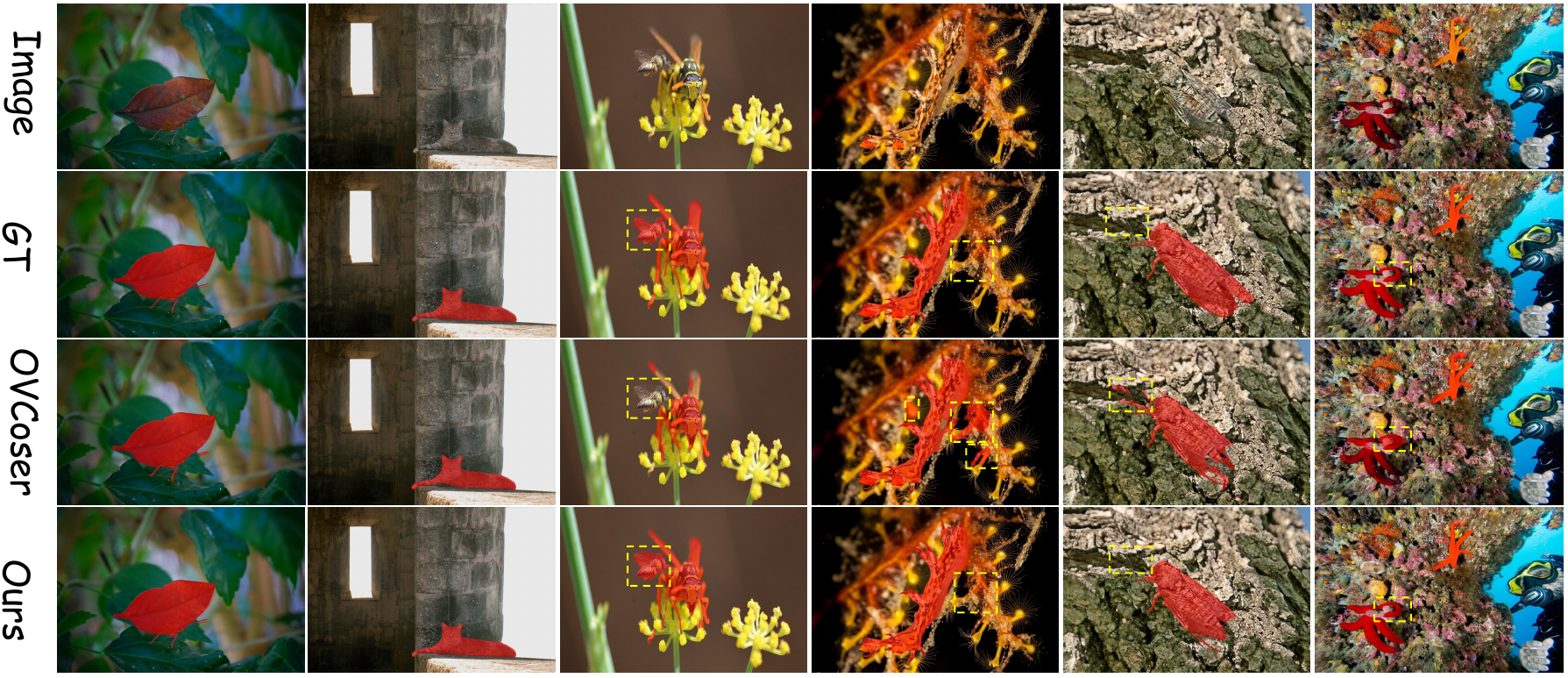}.
    \vspace{-2mm}
    \caption{Qualitative comparison results with the SOTA method OVCoser on the OVCamo dataset.}
    \label{fig:keshihua}
\end{figure}

\parhead{Qualitative Comparison.}
%Fig. \ref{fig:keshihua} presents a qualitative comparison with the SOTA method, OVCoser~\cite{OVCoser}. The visual results highlight OVCoser’s limitations in key areas.
%Qualitative side-by-side mask comparisons further validate our numerical superiority. While existing VLM-based methods suffer from severe semantic drift, often misclassifying background distractors as the target, ViCo-Adapter generates distinct cross-modal anomalies. This results in exceptionally clean boundaries and robust localization, even for rare species embedded in highly cluttered habitats.
\Cref{fig:keshihua} presents a qualitative comparison with the state-of-the-art method OVCoser~\cite{b1}. As shown, OVCoser is more susceptible to background distractors and often produces incomplete or inaccurate masks in highly ambiguous camouflage scenes. In contrast, \ourmethod  ~achieves more accurate localization and cleaner boundaries through improved vision-language alignment, even in complex scenes with rare categories.
\vspace{-2mm}
%\begin{table}[t]
%\caption{Ablation analyses of our proposed modules.}
%\label{tab:ablation}
%\resizebox{\linewidth}{!}{%
%  \renewcommand{\arraystretch}{1.2}%   <--- 调整行距为1.2倍
%  \begin{tabular}{lllllllll}
%    \hline
%    \multicolumn{3}{c|}{\multirow{2}{*}{Method}} & \multicolumn{6}{c}{OVCamo} \\ \cline{4-9}
%    \multicolumn{3}{c|}{}                        & $cMAE\downarrow$ & $cS_m\uparrow$ & $cF_m\uparrow$ & $cF_{\omega}\uparrow$ & $cE_m\uparrow$ & $cIoU\uparrow$ \\ \hline
%    \multicolumn{3}{l|}{Baseline}                & 0.086  & 0.735  & 0.605  & 0.592         & 0.786  & 0.552  \\
%    \multicolumn{3}{l|}{+LoRA}                   & 0.041  & 0.776  & 0.656  & 0.637         & 0.796  & 0.586  \\
%    \multicolumn{3}{l|}{+LoRA + VCDTA}             & 0.017  & 0.883  & 0.854  & 0.835         & 0.940  & 0.767  \\
%    \rowcolor{myzishe}
%    \multicolumn{3}{l|}{+LoRA + VCDTA + BCIM (Ours)}   & \textbf{0.015} & \textbf{0.889} & \textbf{0.859} & \textbf{0.838} & \textbf{0.947} & \textbf{0.773} \\ \hline
%    \multicolumn{3}{l|}{w/o VCDTA + Default prompt}  & 0.019 & 0.881 & 0.837 & 0.822 & 0.939 & 0.761 \\
%    \multicolumn{3}{l|}{w/o VCDTA + No prompt}        & 0.016 & 0.883 & 0.843 & 0.828 & 0.940 & 0.765 \\
%    \multicolumn{3}{l|}{w/o VCDTA + Camouflaged prompt} & 0.020 & 0.871 & 0.846 & 0.822 & 0.935 & 0.744 \\
%    \rowcolor{myzishe}
%    \multicolumn{3}{l|}{Ours}                    & \textbf{0.015} & \textbf{0.889} & \textbf{0.859} & \textbf{0.838} & \textbf{0.947} & \textbf{0.773} \\ \hline
%  \end{tabular}%
%}
%\end{table}
\begin{table}[t]
\caption{Ablation analyses of our proposed modules.}
\label{tab:ablation}
\resizebox{\linewidth}{!}{%
  \setlength{\tabcolsep}{0.6mm}
  \begin{tabular}{lllcccccc}
    \hline
    \multicolumn{3}{c|}{\multirow{2}{*}{Method}} & \multicolumn{6}{c}{OVCamo} \\ \cline{4-9}
    \multicolumn{3}{c|}{}                     & $cS_m \uparrow$ & $cF_\omega \uparrow$ & $cMAE \downarrow$ & $cF_m \uparrow$ & $cE_m \uparrow$ & $cIoU \uparrow$ \\ \hline
    %\multicolumn{3}{l|}{Baseline}             & 0.086  & 0.735  & 0.605  & 0.592         & 0.786  & 0.552  \\
    \multicolumn{3}{l|}{Baseline}& 0.776  & 0.637  & 0.041  & 0.656 & 0.796  & 0.586  \\
    \multicolumn{3}{l|}{Fine-tuned Text Encoder} & 0.837  & 0.774  & 0.029  & 0.786 & 0.883  & 0.695  \\
    \multicolumn{3}{l|}{+ ViCo} & 0.883  & 0.835  & 0.017  & 0.854 & 0.940  & 0.767  \\
    \rowcolor{myzishe}
    \multicolumn{3}{l|}{+ ViCo + ViCoBind}   & \textbf{0.889} & \textbf{0.838} & \textbf{0.015} & \textbf{0.859} & \textbf{0.947} & \textbf{0.773} \\ \hline
    \multicolumn{3}{l|}{w/o ViCo + Default}   & 0.881 & 0.822 & 0.019 & 0.837 & 0.939 & 0.761 \\
    \multicolumn{3}{l|}{w/o ViCo + Class-agnostic}        & 0.883 & 0.828 & 0.016 & 0.843 & 0.940 & 0.765 \\
    \multicolumn{3}{l|}{w/o ViCo + Camouflaged} & 0.871 & 0.822 & 0.020 & 0.846 & 0.935 & 0.744 \\
    \multicolumn{3}{l|}{ViCo + Class-agnostic} & 0.872 & 0.808 & 0.019 & 0.827 & 0.931 & 0.743 \\
    \rowcolor{myzishe}
    \multicolumn{3}{l|}{Ours}                    & \textbf{0.889} & \textbf{0.838} & \textbf{0.015} & \textbf{0.859} & \textbf{0.947} & \textbf{0.773} \\ \hline
  \end{tabular}%
}
\vspace{-3mm}
\end{table}

\begin{figure}[t] 
    \centering
    \includegraphics[width=0.7\columnwidth]{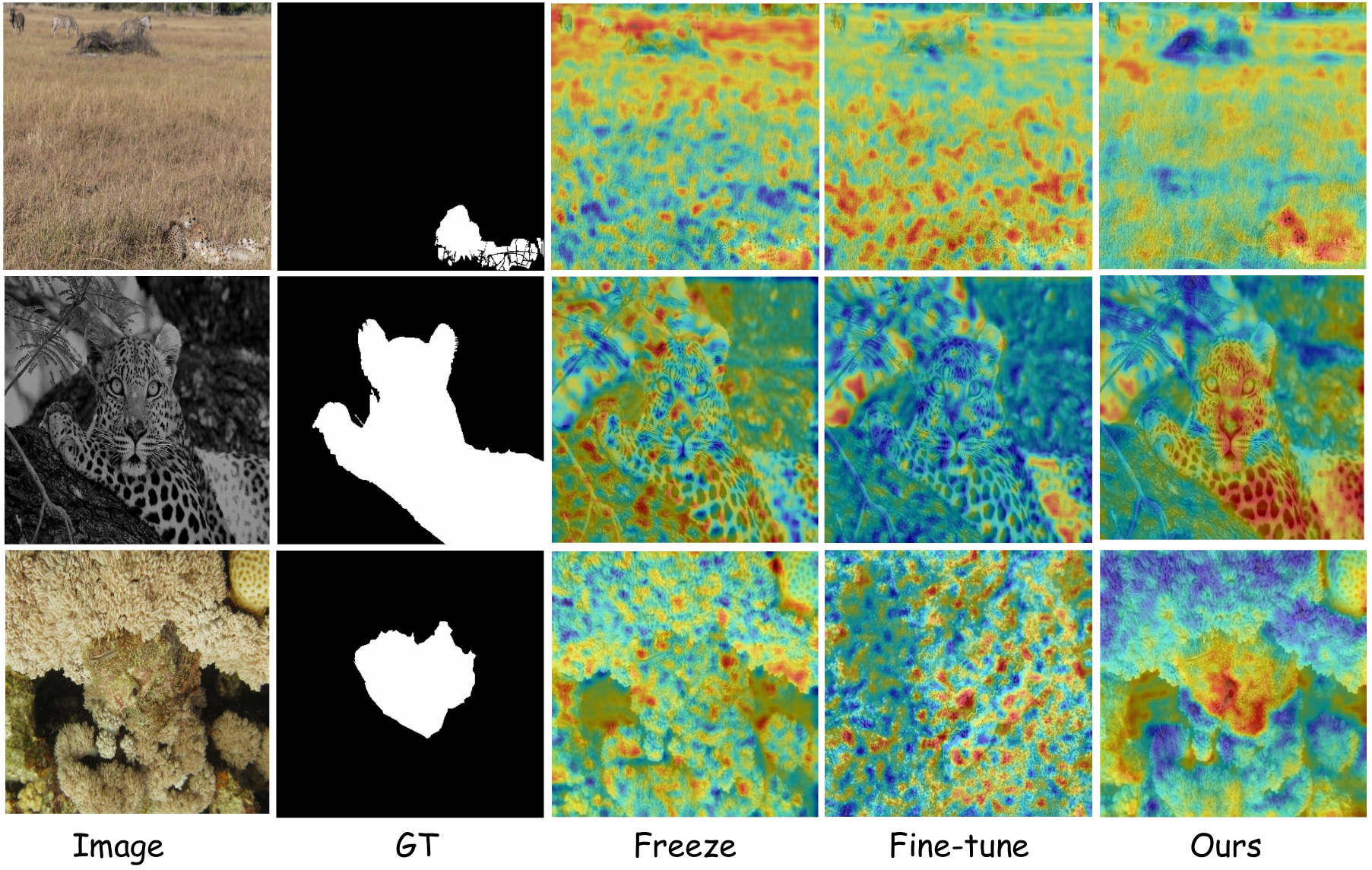}
    \caption{
    Visualization of heat maps using different strategies. From left to right: Input image GT, frozen text encoder, fully fine-tuned text encoder, and our method.
    }
    \label{fig:visual}
    \vspace{-3mm}
\end{figure}
\subsection{Ablation Study}
\label{sec:ablation}
%We conduct extensive ablation studies on the OVCamo dataset to evaluate the core components of our framework and compare our dynamic adaptation against static prompt strategies. Quantitative results are detailed in \textbf{Table \ref{tab:ablation}}.
\parhead{Effectiveness of Key Components.}
%The upper section of \textbf{Table \ref{tab:ablation}} demonstrates the module-wise performance trajectory:
%We use SAM3 as the baseline, keeping the text encoder frozen while allowing the decoder to be trainable. This baseline achieves an IoU of 0.586 but remains limited by the cross-modal semantic gap. Introducing the ViCo substantially improves performance, increasing $cF_m$ from 0.656 to 0.854, demonstrating the effectiveness of dynamically adapting text representations to visual context. Further incorporating the ViCoBind enables bidirectional interaction between visual and textual features, leading to the best overall performance
%, with $cS_m$ of 0.889 and IoU of 0.773. 
%These results verify the complementary roles of dynamic text adaptation and cross-modal binding.
We use SAM3 as the baseline, keeping the encoders frozen while allowing the decoder to be trainable. This baseline achieves an IoU of 0.586 but remains limited by the cross-modal semantic gap. We then unfreeze and fine-tune the text encoder, which improves the IoU to 0.695, indicating that task-specific adaptation of textual representations is beneficial. However, direct fine-tuning still cannot fully bridge the mismatch between textual semantics and image-specific visual cues. Introducing ViCo further improves the IoU to 0.767 and increases $cF_m$ from 0.656 to 0.854, demonstrating the effectiveness of dynamically adapting textual representations to visual context. Further incorporating ViCoBind enables bidirectional interaction between visual and textual features, yielding the best overall performance with a $cS_m$ of 0.889 and an IoU of 0.773.
%\begin{figure}
%    \centering
%    \includegraphics[width=0.8\linewidth]{visual.pdf}
%    \caption{Qualitative comparison results with the SOTA method OVCoser \cite{OVCoser} on the OVCamo dataset.}
%    \label{fig:keshihua}
%   \end{figure}
%\begin{itemize}
%    \item \textit{Baseline:} The frozen SAM3 model struggles with zero-shot camouflage segmentation due to severe visual ambiguity, yielding an $IoU$ of only 0.552.
%    \item \textit{+ LoRA:} Fine-tuning the SAM3 mask decoder with LoRA provides fundamental domain adaptation ($IoU$ improves to 0.586), but fails to bridge the cross-modal semantic gap.
%    \item \textit{+ VCDTA:} Dynamically reshaping text embeddings triggers a massive performance surge. By mitigating semantic confusion, $F_m$ skyrockets from 0.656 to 0.854.
%    \item \textit{+ BCIM (Ours):} The integration of the BCIM module explicitly enables deep bidirectional attention interaction between visual and textual features. This dual-routing fusion pushes the framework to its optimal state, achieving the highest $S_m$ (0.889) and $IoU$ (0.773).
%\end{itemize}

\parhead{Ablation on Prompt Strategies.}
For prompt ablation, we remove ViCo while retaining ViCoBind and feed different static textual representations into ViCoBind. A default class-specific prompt (“a photo of a xxx”), a camouflage-aware prompt, and a class-agnostic prompt (“a photo”) achieve IoUs of 0.761, 0.744, and 0.765, respectively. When ViCo is combined with the class-agnostic prompt, the IoU decreases to 0.743, suggesting that vision-conditioned adaptation still relies on meaningful category semantics as a semantic anchor. Together with the 0.773 IoU achieved by the full model, these results indicate that the gain arises from dynamically adapting category-aware textual representations to image-specific visual evidence rather than prompt engineering alone. As shown in \Cref{fig:visual}, \ourmethod~also produces more concentrated target responses than freezing or directly fine-tuning the text encoder, further demonstrating improved vision-language alignment.
%\textbf{Ablation on Prompt Strategies.} 
%The lower section of \textbf{Table \ref{tab:ablation}} explicitly compares our VCDTA against static prompt designs. 
%\begin{itemize}
%    \item We evaluate a \textit{Default prompt} (i.e., ``a photo of a xxx'') and a \textit{Camouflaged prompt}. Both yield constrained performance (e.g., $IoU$ of 0.761 and 0.744), indicating that rigid linguistic descriptions struggle to represent ambiguous camouflage appearances.
%    \item Interestingly, the \textit{no prompt} setting (using only the general prefix ``a photo'') slightly outperforms specific static prompts ($IoU$ of 0.765). This implies that explicitly naming the class can actually mislead the vision-language model when the target is heavily camouflaged.
%    \item Our full model with VCDTA comprehensively outperforms all static variants by dynamically calibrating the semantic anchor for each image, confirming the absolute superiority of dynamic adaptation over manual prompt engineering.
%\end{itemize}

% \usepackage{multirow}
% \usepackage{multirow}
% \usepackage{multirow}

\section{Conclusion}
\label{sec:conclusion}
\vspace{-1mm}
We present \ourmethod ~for open-vocabulary camouflaged object segmentation (OVCOS). Motivated by the semantic mismatch between textual representations and fine-grained camouflage cues, we introduce a vision-conditioned (ViCo) module to dynamically adapt text features using image-specific visual context, while preserving the open-vocabulary knowledge inherited from pre-training. We further design a vision-conditioned cross-modal binding module (ViCoBind) to enhance bidirectional vision-text interaction and fine-grained semantic alignment. Experiments on OVCamo demonstrate SOTA performance and strong generalization to rare and unseen categories. 
We hope this work will facilitate the application of foundation models to OVCOS.
%In this paper, we present \textbf{\ourmethod}, a novel parameter-efficient framework that adapts the Segment Anything Model 3 (SAM3) for Open-Vocabulary Camouflaged Object Segmentation (OVCOS). To overcome the severe semantic drift bottleneck caused by rigid text prompts in highly ambiguous environments, we introduce the Vision-Conditioned Dynamic Text Adaptation (VCDTA) mechanism. By actively extracting environmental visual priors to dynamically reshape textual anchors via a gated residual offset, our method effectively transforms camouflaged targets into distinct cross-modal anomalies without disrupting the foundation model's pre-trained knowledge. Coupled with lightweight sequential visual adapters and deep bidirectional feature interaction, ViCo-Adapter establishes a new state-of-the-art on the OVCamo benchmark. Extensive experiments demonstrate that our vision-conditioned paradigm provides a highly robust, elegant, and efficient solution for adapting large foundation models to complex camouflaged scenarios, maintaining exceptional zero-shot generalization on unseen and rare species.
%\vfill\pagebreak

\fontsize{9pt}{11pt}\selectfont

\bibliographystyle{IEEEbib}
% \bibliography{strings,refs}
\bibliography{refs}
\end{document}